\documentstyle[colap, epsfig]{article}
\input{epsf}

\newcommand{\clig}{{\sl CLiG} }
\newcommand{\clear}{{\sc Clears} }

\title{An Education and Research Tool for Computational Semantics}
\author{Karsten Konrad, Holger Maier \and Manfred Pinkal \\
Computerlinguistik, Universit\"at des Saarlandes \\ 66041 Saarbr\"ucken, Germany \\ konrad, maier, pinkal@coli.uni-sb.de \\[4mm]
\And David Milward \\
SRI International, \\ Suite 23, Millers Yard \\ Cambridge, CB2 1RQ, GB \\
milward@cam.sri.com}

\begin{document}
\maketitle

\section*{Abstract}
This paper describes an interactive graphical environment for
computational semantics. The system provides a teaching tool, a stand alone
extendible grapher, and a library of
algorithms together with test suites. The teaching tool allows
users to investigate the properties of various semantic formalisms
(e.g. Intensional Logic, DRT, and Situation Semantics), and their interaction with
syntax.

\section{Introduction}

The \clear tool (Computational Linguistics Education and Research Tool
in Semantics) was developed as part of the FraCaS project, LRE 62-05,
which aimed to encourage convergence between different semantic
formalisms. Although formalisms such as `Montague-Grammar', DRT, and
Situation Semantics look different on first sight, they share many
common assumptions, and provide similar treatments of many phenomena.
The \clear tool allows exploration and comparison of these different
formalisms, enabling the user to get an idea of the range of
possibilities of semantic construction.  It is intended to be used as
both a research tool and a tutorial tool.

The first part of the paper shows the potential of the system for
investigating the properties of different semantic formalisms, and for
teaching students formal semantics.  The next section outlines the
library contents and the system architecture, which was designed to
reflect convergence between theories. The result is a highly modular
and, we believe, a highly flexible system which allows user programs
to be integrated at various levels.  The final part of the paper
describes the grapher which was designed as a stand alone tool which
can be used by various applications.

\section{A Tutorial System for Computational Semantics}

As a tutorial tool, \clear allows students to investigate by themselves certain formalisms and
their relationship. It also provides the possibility for the teacher to provide
interactive demonstrations and to produce example slides and handouts.

In this section we will show how a user could interactively explore
the step-by-step construction of a semantic representation out of a syntax tree.
The following figures show the initial display for the sentence ``Anna laughs''
in `Montague Grammar' \cite{DowWalPet:81} and compositional versions of DRT
\footnote{$\otimes$ is the symmetric {\tt merge} operation in $\lambda$-DRT,
which puts together two DRSs by unioning the discourse universes and the conditions.}
($\lambda$-DRT \cite{BMMMP:94} and C-DRT of \cite{Muskens:93}).

\begin{figure}
\noindent
\begin{minipage}[b]{\linewidth}
  \centering
  \epsfig{figure=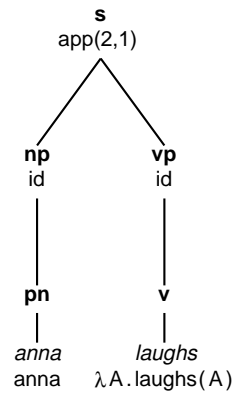}
  \caption{Initial Representation of {\tt Anna laughs} with `Montague Grammar'}
  \label{fig:lgq1}
\end{minipage}\hfill
\end{figure}

\begin{figure}
\noindent
\begin{minipage}[b]{\linewidth}
  \centering
  \epsfig{figure=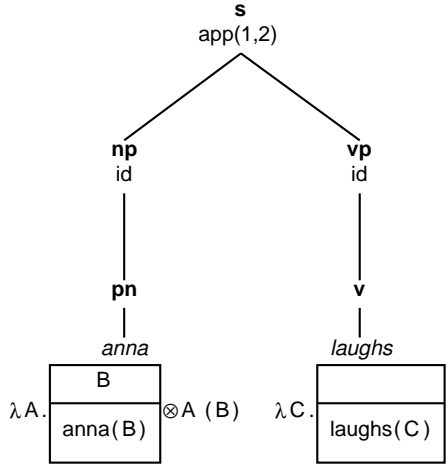}
  \caption{Initial Representation of {\tt Anna laughs} with $\lambda$-DRT}
  \label{fig:ldrt1}
\end{minipage}\hfill
\end{figure}

\begin{figure}
\noindent
\begin{minipage}[b]{\linewidth}
  \centering
  \epsfig{figure=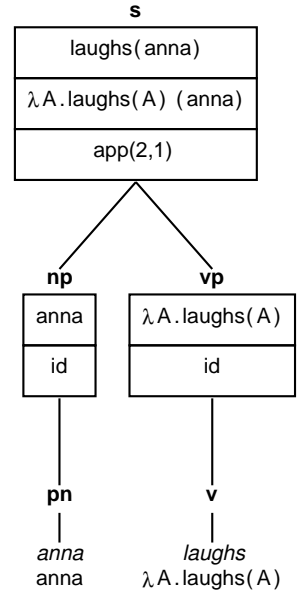}
  \caption{Final Representation of {\tt Anna laughs} in `Montague-Grammar'}
  \label{fig:lgq3}
\end{minipage}\hfill
\end{figure}

\begin{figure}
\noindent
\begin{minipage}[b]{\linewidth}
  \centering
  \epsfig{figure=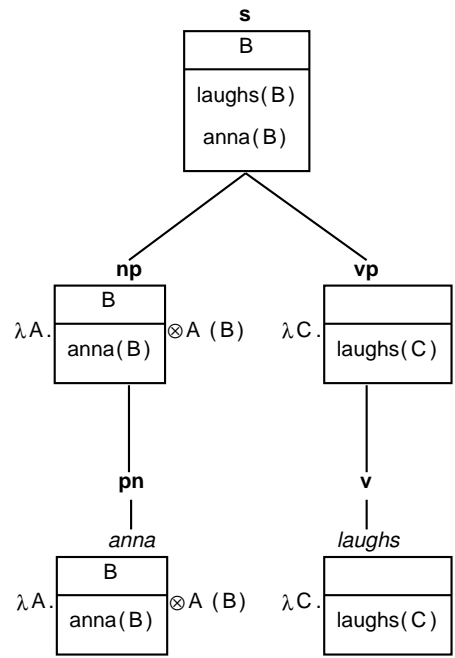}
  \caption{Final Representation of {\tt Anna laughs} in $\lambda$-DRT}
  \label{fig:ldrt2}
\end{minipage}\hfill
\end{figure}

The user controls a derivation by moving to particular nodes in the
derivation tree and by using either mouse clicks or a pull-down
menu. The menu provides the option of fully processing the current
node (and the nodes below it), or performing single derivation steps
(e.g. intensional operator cancellation, quantifier storage, and DRS
merging).  Figures \ref{fig:lgq3} and \ref{fig:ldrt2} show possible
final representations for the examples above.  The different graphical
outputs shown here are controlled by the user via parameter setting.

 
\section{Comparing theories}

\begin{figure}
\noindent
\begin{minipage}[b]{\linewidth}
  \centering
  \epsfig{figure=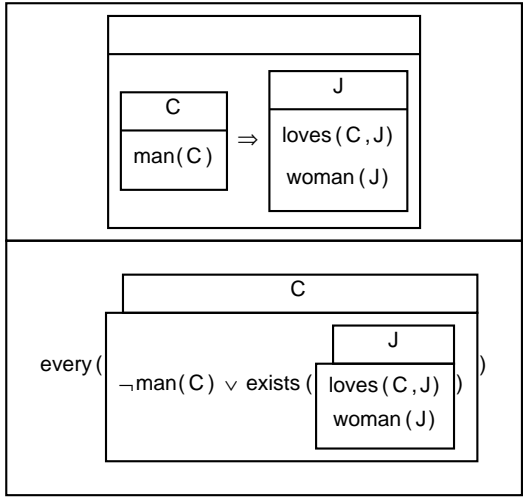}
  \caption{Translation from DRT to Predicate Logic}
  \label{fig:trans1}
\end{minipage}\hfill
\end{figure}

A major use of the tool is for comparison of different semantic
theories. To this end facilities are provided for simultaneously
building up representations using different formalisms. Furthermore
there are translation routines among some semantic formalisms, making
easy comparison of various results possible. In figure \ref{fig:trans1}
we show a translation from a DRS to a formula in Predicate Logic.
As our system is controlled by parameters, controlling the choice
of semantic formalism, parser, grammar, syntax-semantics mapping etc. 
the user can also try out several thousand different parameter 
settings and comparing their different results.

\section{The Library}

Because such a tutorial system as described above has to be based
largely on standard routines and algorithms that are fundamental for
the area of computational semantics, a secondary aim of the project
was to provide a set of well documented programs which could form the
nucleus of a larger library of reusable code for this field. For
program documentation we largely followed the approach taken in LEDA
\cite{Naher:93}).

\subsection{Tools for Semantic Construction} The library currently
contains a selection of lambda reduction routines (including routines
based on unification and on substitution plus alpha-conversion),
routines for performing functional composition, routines for
performing quantifier storage (Cooper storage \cite{Cooper:83} and 
nested Cooper storage \cite{Keller:88}), a simple and a 
feature based phrase structure grammar and a
DCG parser, a feature based Categorial Grammar with an incremental
parser, a couple of syntactic lexicons, various sets of lexical macros
for different semantic formalisms (currently including Compositional
DRT \cite{Muskens:93}, $\lambda$-DRT \cite{BMMMP:94}, Logic of
Generalised Quantifiers, and Intensional Logic \cite{DowWalPet:81}),  
as well as translation routines from representations in DRT to 
formula in Predicate Logic or Situation Semantics.
There are also
modules for mapping from syntax trees to semantic representations
using rule-to-rule based mappings, or mapping via syntactic templates.

The routines concerned with the interface to the grapher \clig
consist of translation routines from the Prolog representations of the
semantic formalisms to the \clig syntax and parameters to manipulate
the display format, such as stacking of reduction steps or drawin boxes 
around specific nodes, as you can see in figure \ref{fig:ldrt2}. 
It is intended to extend the library with routines for semantic 
construction driven by semantic types, and to integrate a wider
range of grammars, parsing strategies and pronoun resolution
strategies.

\subsection{Modularisation Principles}

One of the aims in building the tool was to show where semantic
formalisms converge. Thus there was theoretical motivation as well as
practical motivation in making sure components of the system were
shared wherever possible. Achieving this was not entirely trivial
however, since many of the changes required in moving between
formalisms cut across natural modularisation boundaries. For example,
use of intensional logic rather than first order logic requires a
change in the quantifier discharging rules, and even in the lambda
reducer (the natural way of reducing an intensional expression
involves interleaving of beta-reductions with intensional operator
cancellation).

The solution adopted was to use {\it parameterised modularisation}.
Whenever some part of the code needed to be different, a parameterised
level was introduced at that point. At run time, the parameter setting
chooses the correct module.

The parameterised approach involves some small costs due to
indirection (instead of calling e.g. a $\beta$-reducer directly, a
program first calls a routine which chooses the $\beta$-reducer
according to the parameters).  But with these parameterisation layers
we provide natural points where the system can be extended or modified
by the user.  The approach also prevents `generalisation to the worst
case' which is often seen in other approaches to modularisation. For
example, the parser calls a parameterised level which chooses how to
annotate nodes, so that the syntax trees only have the information
required for the particular syntax-semantics strategy being used.
The result is a system which provides several thousand possible valid
combinations of semantic formalisms, grammer, reducer etc. using a
small amount of code.

\section{The graphical interface}

\begin{figure}
\noindent
\begin{minipage}[b]{\linewidth}
  \centering
  \epsfig{figure=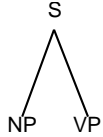}
  \caption{Tree defined by the above description string}
  \label{fig:tree}
\end{minipage}\hfill
\end{figure}

Another major part of our work on the educational tool was the development of
a general graphical browser or {\it grapher\/} for the graphical
notations used in computational linguistics, especially those in
computational semantics such as trees, Attribute-Value-Matrices, EKN
\cite{BarCoo:93} and DRSs.
Two other design points which were especially important were the {\it
extendibility\/} of the grapher for future applications and the
possibility of {\it interaction\/} between the user, the grapher and
the underlying application.
The grapher was written in Tcl/Tk, a programming system for developing
graphical user interfaces (see \cite{Ousterhout:94}). This system
has the further advantage of providing a translation routine from 
graphic canvases into Postscript. 
Using this feature way we produced all figures in this paper 
with the \clear tool.
Graphical structures are described using a {\sl description string},
a plain text hierarchical description of the object to be drawn 
without any exact positioning information. The next example describes a 
simple tree consisting of a mother node S and the two daughter nodes
NP and VP. The graphical result of this description is shown in 
figure \ref{fig:tree}.

\begin{itemize}
\item \begin{verbatim}
   {tree 
      {plain-text "S"} 
         {plain-text "NP"} 
         {plain-text "VP"}}

      \end{verbatim}
\end{itemize}

Furthermore {\sc Clig} can display {\it interactive\/} graphical structures
which allow the user to perform actions by clicking on mouse-sensitive
regions in the display area. The grapher and an underlying application
therefore can behave in a way that the grapher is not only a way to
visualise the data of the application, but also could be seen as a real
interface between user and application.

The mode of interactivity is totally under the control of the
application currently using the grapher.

\section{Availability of the system}

The system is available at the ftp address:  {\tt ftp.coli.uni-sb.de:/pub/fracas}
or on the WWW at the URL {\tt http://coli.uni-sb.de/~maier/clears.html}.
For further and more detailed information on the results of the FraCaS project, you should read 
\cite{fracas:d15} or \cite{fracas:d16}.
These papers are electronically available at the same ftp-directory as the \clear tool.
You should also have a look at the FraCaS homepage on the WWW at URL {\tt 
http://www.cogsci.ed.ac.uk:80/~fracas/}.

\section{Conclusion}
Initial reactions to demonstrations of the educational tool
suggest that it has the potential to become a widely used
educational aid. We also believe that the programs implemented and
documented in this work provide the nucleus
of a larger library of reusable programs for computational semantics.
Our current plans are to test the system with a wide class of users
to discover areas requiring extension or modification. A longer term
aim would be to integrate the system with existing grammar development
environments.

%------------------
\bibliographystyle{acl}
\bibliography{short}

% NOTE HOLGER HAD BIBLIOGRAPHY KONVENS
\end{document}